\documentclass[journal]{IEEEtran}

\usepackage{mathtools}
\usepackage{multirow}
\usepackage{pythonhighlight}
\usepackage{float}
\usepackage[caption=false]{subfig}
\usepackage{makecell}
\usepackage{booktabs}
\usepackage{amsmath,amsfonts}
\usepackage{algorithmic}
\usepackage{algorithm}

\usepackage{times}
\usepackage{epsfig}
\usepackage{amssymb}
\usepackage[absolute]{textpos}
\makeatletter
\let\NAT@parse\undefined
\makeatother
\usepackage{hyperref}
\newfloat{figtab}{t}{t}
\makeatletter
\newcommand\figcaption{\def\@captype{figure}\caption}
\newcommand\tabcaption{\def\@captype{table}\caption}
\makeatother

\begin{document}
%
\title{Dissecting RGB-D Learning for Improved Multi-modal Fusion}
%
%

\author{Hao Chen, Haoran Zhou, Yunshu Zhang, Zheng Lin, Yongjian Deng
\thanks{Manuscript received April 19, 2021; revised August 16, 2021.
Hao Chen, Haoran Zhou, Yunshu Zhang are with the School of Computer Science and Engineering, Southeast University, Nanjing, China (e-mail: haochen303@seu.edu.cn; haoranzhou0@gmail.com; zhangyunshu@seu.edu.cn). They are also with Key Laboratory of New Generation Artificial Intelligence Technology and Its Interdisciplinary Applications (Southeast University), Ministry of Education, China. 
Zheng Lin is with Tsinghua University (frazer.linzheng@gmail.com) and Yongjian Deng is with the College of Computer Science, Beijing University of Technology (yjdeng@bjut.edu.cn).
}}

\maketitle
\begin{abstract}

In the RGB-D vision community, extensive research has been focused on designing multi-modal learning strategies and fusion structures. However, the complementary and fusion mechanisms in RGB-D models remain a black box. In this paper, we present an analytical framework and a novel score to dissect the RGB-D vision community. Our approach involves measuring proposed semantic variance and feature similarity across modalities and levels, conducting visual and quantitative analyzes on multi-modal learning through comprehensive experiments. Specifically, we investigate the consistency and specialty of features across modalities, evolution rules within each modality, and the collaboration logic used when optimizing a RGB-D model. Our studies reveal/verify several important findings, such as the discrepancy in cross-modal features and the hybrid multi-modal cooperation rule, which highlights consistency and specialty simultaneously for complementary inference. We also showcase the versatility of the proposed RGB-D dissection method and introduce a straightforward fusion strategy based on our findings, which delivers significant enhancements across various tasks and even other multi-modal data. 
\end{abstract}

\section{Introduction}
\label{sec:intro}

With the advent of new visual sensors, multi-modal visual systems such as smartphones \cite{kelly2020multimodal}, mobile robots \cite{hu2018small}, virtual reality equipments \cite{wang2019multimodal}, and autonomous driving systems \cite{feng2020deep} have gained immense popularity. 
The aim of these systems is to integrate complementary cues from different visual modalities, thereby improving the accuracy and robustness of visual perception and understanding. To achieve this goal, diverse multi-modal vision models \cite{Refinenet,RDF,PCA,R9,R6,R11,DMRA} have been proposed, particularly those based on deep neural networks. These models have achieved significant improvements compared to their unimodal counterparts.
\par However, the mainstream philosophy in this domain is to design various data-driven cross-modal interaction strategies \cite{Cat-det,MIM,MM-TTA,zhao2022modeling,Deepfusion} and combination paths \cite{shvetsova2022everything,Refinenet,RDF,PCA,R9,R6,R11,DMRA} to enhance fusion sufficiency. These strategies are often based on heuristic hypotheses or empirical inheritance. 

\begin{figure}[htbp]
\begin{center}
   \includegraphics[width=\linewidth]{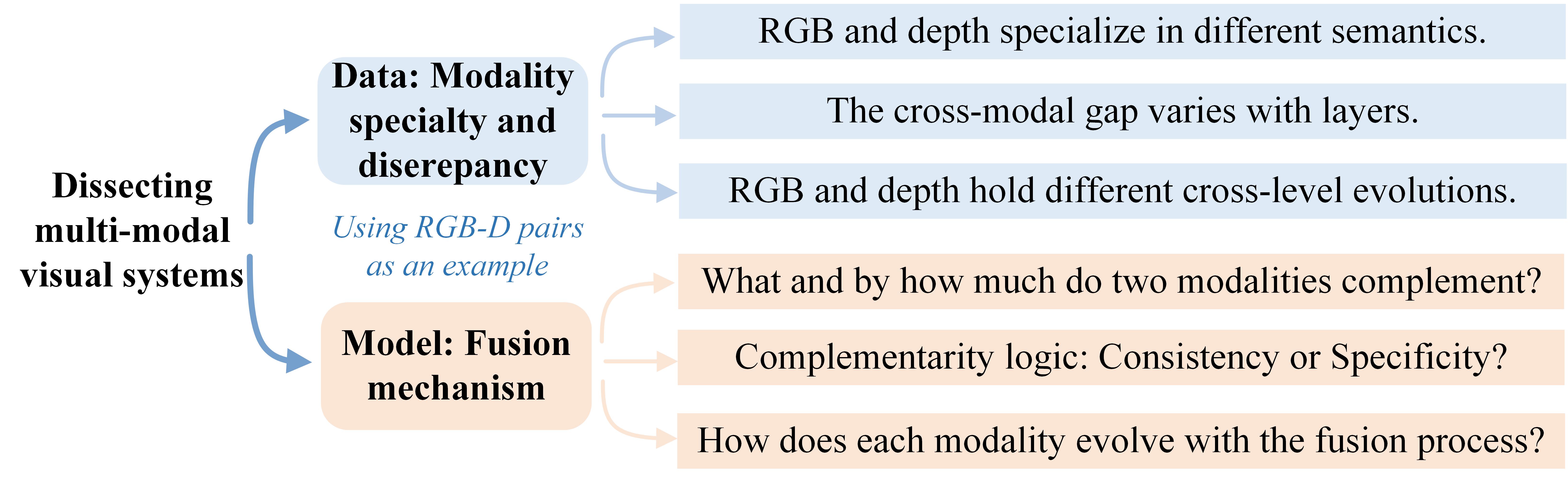}
\end{center}
   \caption{
   Our motivation and analytical structure of the dissection on RGB-D visual fusion.}
\label{fig:introduction}
\end{figure}

\par For instance, a two-stream architecture \cite{Refinenet,RDF,PCA,R9,R6,R11,DMRA} often symmetrically integrates cross-modal features as the default fusion choice. However, with distinguished modality specialties and severe cross-modal discrepancies, such layer-by-layer cross-modal alignment is actually difficult to establish. Another example is the debate on the role of cross-modal consistency in improving fusion performance. Studies advocating minimizing \cite{MIM} and maximizing \cite{Mmss} cross-modal consistency are both widely popular without consensus. 
Although their task-specific improvements are experimentally verified, the lack of theoretical explanations makes it difficult to rule out the possibility of overfitting and limits their generalizability.

\par Instead, by being aware of the concrete distribution of cross-modal complements and the complementing mechanism in the fusion process, we can design a RGB-D model with fewer parameters, better accuracy, and more trustworthy results, while carrying better generalization ability to other tasks and multi-modal pairs. 
To fill this gap, we here make the first attempt to dissect the complementarity and fusion mechanism of RGB-D data. As illustrated in Figure~\ref{fig:introduction}, our motivation encompasses two key aspects:  (a) dissecting the distinct characteristics and disparities in semantic distribution and cross-level evolution across modalities, and (b) unraveling the multi-modal fusion mechanism. This includes quantifying the cross-modal semantic complementarity, examining the logic behind how it promotes modal consistency or specificity, and investigating the evolution of both modalities through multi-modal learning. 
\par We achieve these by proposing the multi-modal dissection framework, which features a dissection architecture for simultaneous analysis of multi-modal data complementarity and fusion mechanisms, a new evaluation score called ``semantic variance" to quantify concept-level semantic discrepancy between distributions, and a fusion strategy derived from dissection findings to improve efficiency.
Our dissection framework for RGB-D fusion is shown in Figure~\ref{fig:framework interpretation}. 
To comprehensively expose the multi-scale cross-modal complements, we adopt semantic segmentation \cite{song2015sun}, a dense prediction task that requires multi-scale cues, to perform our study. 

To investigate the modality difference in cross-level evolution and explore whether cross-modal feature symmetry holds, we propose the Cross-modal Symmetry Measurement (CSM) module, where cross-layer and cross-modal feature similarities are computed and compared. We also introduce the Semantic Distribution Contrast (SDC) module and the semantic variance score to quantify the semantic discrepancy between the modalities. SDC maps high-level features with Net2Vec \cite{fong2018net2vec} to obtain semantic distributions for each modality, which are then compared to analyze the discrepancy in each semantic concept. 
The proposed semantic variance score weighted summarizes such concept-wise discrepancy by differentiating the significance between concept fluctuation and occurrence. By comparing the semantic discrepancy before and after multi-modal fusion, we can clearly dissect the cross-modal fusion gains, and the evolution of each modality during the fusion process. Based on these analyses, we find that the multi-modal fusion logic is a hybrid one that highlights cross-modal consistency and specificity simultaneously. We hence design a simple yet effective strategy that minimizes and maximizes cross-modal mutual information to boost multi-modal fusion efficiency. This strategy is parameter-free and plug-and-play for different tasks and multi-modal data.

In summary, our contributions include: 

1) To the best of our knowledge, this study, for the first time, systematically analyzes RGB-D fusion and introduces a comprehensive multi-modal dissection framework to the field.

2) Designing a score to measure inter-modal semantic discrepancy and semantic gains with cross-modal fusion.

3) Conducting comprehensive experiments to reveal/verify important insights in RGB-D fusion, and based on the findings, proposing an effective fusion strategy to improve multi-modal fusion sufficiency. Further experiments demonstrate the versatility of our multi-modal dissection framework and fusion strategy.

\begin{figure*}[ht]
\begin{center}
   \includegraphics[width=\linewidth]{figures/framework-interpretation_final.jpg}
\end{center}
    \vspace{-10pt}
   \caption{Dissection framework for RGB-D fusion. (a) The main pipeline of our framework. (b) The process of Cross-modal Symmetry Measurement, which calculates the similarity between cross-level and cross-modal features with CKA (\ref{sec:3.2.2}). (c) The process of Semantic Distribution Contrast (SDC), which obtains the semantic distribution in features with Net2Vec~\cite{fong2018net2vec}, contrasts their semantic distribution, and calculates the SVar (\ref{sec:3.2.1}) score.}
\label{fig:framework interpretation}
\end{figure*}

\section{Related Work}
\label{sec:rela}

\subsection{Multi-modal visual systems}
Most existing multi-modal visual systems aim to improve the cross-modal fusion efficiency for different modalities
by enhancing the cross-modal interaction paths from single-level early \cite{DeepFuse}/late \cite{Localsensi,Mmss} fusion to diverse multi-level ones \cite{Refinenet,RDF,PCA,R9,R6,R11,DMRA} with a two-stream architecture. Following works widely adopt
attention mechanism (e.g., \cite{Deepfusion,TANet}) or append another joint encoding branch \cite{zhou2021specificity,TANet} to enhance the fusion sufficiency. 
However, these fusion patterns may have redundant parameters and overfitting risks due to their complexity.  
\par Apart from crafting the fusion paths, another line of works introduces feature-level or inference-level constraints to promote mutually complementary optimization, such as maximizing cross-modal feature speciality \cite{Localsensi,Discri} or consistency \cite{Mmss}, minimizing cross-modal mutual information \cite{MIM}, cross-modal contrastive augmentation \cite{Cat-det}, cross-modal mutual supervision \cite{MM-TTA} and  cross-modal distillation \cite{CMD,LSF}. 
Differently, our motivation is to dissect the cross-modal complementarity/consistency and fusion logic with comprehensive experimental explanations, rather than boost task-specific accuracy with heuristic or empirical assumptions. 

\subsection{Deep Neural Network Interpretation}


Visualizing the visual concepts encoded in the intermediate layers of DNNs is a natural way to interpret DNNs. Activation maximization \cite{erhan2009visualizing, wang2018visualizing, mahendran2015understanding, yosinski2015understanding, nguyen2016synthesizing, olah2017feature} finds a representative feature that maximizes activation of a neuron, channel or layer. People can roughly decode visual concepts contained in intermediary layers of DNNs using visualization results. 
\par Besides visualization, some studies have explored the relationships between filters and visual concepts (e.g., materials, certain objects). Bau et al.~\cite{bau2017network} collect a dataset with pixel-wise labels of visual concepts, and find the alignment between a single filter and a specific concept. Fong and Vedaldi \cite{fong2018net2vec} show how a DNN uses multiple filters to represent a particular semantic concept, and investigate the embeddings of concepts with a combination of several filters.

\par Another way to explain DNNs is using Game-theoretic methods.  Several studies have explored the interactions among multiple input variables of the DNN~\cite{ren2024we, chen2024defining, zhou2024explaining, deng2024unifying, zhou2024interpretability, ren2023defining, liu2023towards} with the Shapley value~\cite{shapley1953value}. Additionally, some research has investigated the cross-modal interactions with the Shapley value~\cite{li2023boosting, wei2024enhancing} and the Banzhaf interaction~\cite{jin2023video}.

\par \textit{Difference.} To summarize, most existing interpretation works only explore the internal mechanisms of unimodal models, while the analysis of the multi-modal data complementarity and multi-modal fusion mechanism is still lacking. To our knowledge, MultiViz~\cite{liang2022multiviz}, \cite{wang2020makes} and \cite{huang2021makes} are the closest related works. However, they focus on the visualization of cross-modal interactions and assigning concepts to decision-level features \cite{liang2022multiviz}, the potential performance degradation in multi-modal systems, by examining the gradient imbalance from the optimization perspective \cite{wang2020makes}, and theoretically discussing the latent representation bound from the generalization perspective \cite{huang2021makes}. In contrast, our work has different motivations and core problems. Our main motivation is to dissect cross-modal relation and multi-modal learning mechanism, thus providing a systematic analysis to benefit the RGB-D and even other multi-modal communities. Based on this, we study the cross-modal data consistency and discrepancy, fusion logic, and modality evolution rules with semantic and quantitative explanations. From the application perspective, our study and conclusions can not only help improve the performance of a multi-modal visual system, but also provide guidance to diagnose a multi-modal system and design a universally efficient multi-modal architecture for various tasks and modalities.



\section{Method}
\label{sec:3_method}

\subsection{Overview}
\par In Figure~\ref{fig:framework interpretation}, we illustrate the process of our framework. Initially, we train separate unimodal networks to obtain encoders and features for RGB and depth, denoted as ${E_I^S, F_I^S}$ and ${E_D^S, F_D^S}$, respectively. Subsequently, we employ a simple two-stream late fusion architecture to jointly train both modalities. This architecture fuses the features from both modalities to create the joint RGB-D feature $F_C^J$, enabling collaborative inference. The encoders and features learned through joint training are denoted as ${E_I^J, F_I^J}$ and ${E_D^J, F_D^J}$ for RGB and depth, respectively. After training, we obtain the semantic distribution of each feature and compare them to analyze the semantic discrepancy among features using Net2Vec~\cite{fong2018net2vec}. To quantify the semantic variance between features, we utilize our designed semantic variance score. Additionally, we measure feature similarity across all layers using the Linear Centered Kernel Alignment (CKA)~\cite{kornblith2019similarity}. For qualitative analysis, we generate the saliency mask corresponding to each concept using Grad-CAM~\cite{selvaraju2017grad}.

\par To dissect the complementarity of the cross-modal data, we compare the characteristics between the two modalities. For dissecting multi-modal fusion logic, we evaluate the disparity among the separately trained unimodal features, the jointly trained counterparts, and the fusion feature using our framework.


\subsection{Metrics}

\subsubsection{Semantic Variance}
\label{sec:3.2.1}
\par We analyze the distribution of semantic concepts encoded by each unimodal model and the multi-modal fusion model using the Set IoU score based on Net2Vec \cite{fong2018net2vec}.

\noindent \textbf{Preliminaries: Net2Vec.} Fong and Vedaldi \cite{fong2018net2vec} proposed Net2Vec, a method to investigate semantic concepts in DNNs by aligning them to filters. Net2Vec records filter activations when probed with a reference dataset and learns to weight them for semantic tasks.

As we here use the semantic segmentation task as an example to dissect the RGB-D fusion process, we only describe the method for learning segmentation concept embeddings. To this end, Net2Vec learns weights $\mathbf{w} \in \mathbb{R}^K$, where $K$ is the number of filters in the target layer, to linearly combine thresholded activations. The sigmoid function $\sigma$ is then applied to the linear combination to predict a segmentation mask $M(\mathbf{x} ; \mathbf{w})$:

\begin{equation}
M(\mathbf{x} ; \mathbf{w})=\sigma (\sum_k w_k \cdot \mathbb{I}\left(A_k(\mathbf{x})>T_k\right)  ) ,
\end{equation}
where $\mathbb{I}(\cdot)$ is the indicator function of an event, $A_k(\mathbf{x})$ is the activation map of filter $k$ on input $x$, and $T_k$ is the threshold of filter $k$ that determines the highlighting area of the target concept. Note that, in this paper, we do not threshold the activations to more precisely assess the semantic distribution. During training, the weights $w$ of concept $c$ are learned by minimizing a per-pixel binary cross-entropy loss weighted by the mean concept size:

\begin{equation}
\begin{aligned}
\mathcal{L}_1=-\frac{1}{N_{c}} & \sum_{\mathbf{x} \in X_{c}} \alpha M(\mathbf{x} ; \mathbf{w}) L_c(\mathbf{x}) \\
& +(1-\alpha)\left(1-M(\mathbf{x} ; \mathbf{w})\left(1-L_c(\mathbf{x})\right)\right. ,
\end{aligned}
\end{equation}
where ${N_{c}}$ is the number of samples that contain the concept $c$, the symbol $\mathbf{x} \in X_{c}$ denotes the samples that contain concept $c$, and $\alpha = 1-\sum_{x \in X_{train}} \left|L_c(x)\right|/S$, where $\left|L_c(x)\right|$ is the number of foreground pixels for concept $c$ in the ground truth mask for $x$ and $S=h_s \cdot w_s$ is the number of pixels in ground truth masks.

After training, we calculate the Set IoU score for each concept $c$ to demonstrate the semantic distribution encoded in a layer. The score is given by:

\begin{equation}
IoU_{set}\left(c ; M\right)=\frac{\sum_{\mathbf{x} \in X_{c}}\left|M(\mathbf{x}) \cap L_c(\mathbf{x})\right|}{\sum_{\mathbf{x} \in X_{c}}\left|M(\mathbf{x}) \cup L_c(\mathbf{x})\right|} ,
\end{equation}
which calculates the data-set-wide intersection over union score (Jaccard index) between the ground truth segmentation masks $L_c$ and the binary segmentation masks $M$ predicted by the linear combination of activations with weights $\mathbf{w}$.




\noindent \textbf{Semantic Variance.} Given the features from two networks, $F_1$ and $F_2$, we use the symbol ${IoU_{set}(c_j;F_i)}_{j=1...C}$, where $C$ is the number of concepts, to denote the distribution of Set IoU in feature $F_i$. To quantify the semantic gains of $F_2$ in comparison to $F_1$, we use ${SVar}_{c_j}^1\left(F_2 ; F_1\right)$:
\begin{equation}
{SVar}_{c_j}^1\left(F_2 ; F_1\right)=\frac{IoU_{set}(c_j;F_2)-IoU_{set}(c_j;F_1)}{max(IoU_{set}(c_j;F_2),IoU_{set}(c_j;F_1))} ,
\end{equation}
to represent the change between $F_1$ and $F_2$, which measures the score of semantic gains over $F_1$ for concept $j$. We consider that generating a new concept is typically more challenging than increasing the Set IoU values for existing concepts in the feature. Therefore, we assign a higher score to new concepts compared to other concepts that experience the same change but already exist in $F_1$. Similarly, we give greater weight to the score of disappeared concepts. To assess the semantic variance for each new or disappeared concept, we utilize the average Set IoU across all concepts as a baseline. The semantic variance is given by:

\begin{equation}
{SVar}_{c_j}^2\left(F_2 ; F_1\right)=\frac{IoU_{set}(c_j;F_2)-IoU_{set}(c_j;F_1)}{\frac{1}{C}\sum_{k=1}^C IoU_{set}(c_k;F_1)} ,
\end{equation}

To address the potential bias introduced by the varying proportions of concepts in the dataset, which may result in the network paying more attention to concepts with larger sample sizes, we normalize the scores of all concepts by dividing them by their respective proportions in the dataset. Based on these insights, we calculate a scalar semantic variance by summing the scores for each concept. This enables us to quantify the overall semantic improvements of $F_2$ compared to $F_1$. We formulate the semantic variance between $F_1$ and $F_2$ as follow:

\begin{equation}
{SVar}\left(F_2 ; F_1\right)=\sum_{j=1}^c \frac{\alpha {SVar}_{c_j}^1+\lambda(1-\alpha) {SVar}_{c_j}^2}{\frac{\sum_{x \in X_R}\left|L_j(x)\right|}{\left|X_R\right| S}} ,
\end{equation}

\noindent where $\alpha = \mathbb{I}(min(IoU_{set}(c_j;F_2),IoU_{set}(c_j;F_1))>0)$, $X_R$ is the reference dataset, $\left|L_j(x)\right|$ is the number of foreground pixels for concept $j$ in the ground truth mask for $x$, $\left|X_R\right|$ is the number of samples in the reference dataset, and $S=h_s \cdot w_s$ is the number of pixels in ground truth masks. ${SVar}_{c_j}^1\left(F_2 ; F_1\right)$ denotes the semantic variances for increased/decreased concepts, while ${SVar}_{c_j}^2\left(F_2 ; F_1\right)$ represents the semantic variances for newly emerging/vanishing ones. $\lambda$ is a weight used to emphasize newly emerging/vanishing concepts and is set to 2. A positive semantic variance indicates that $F_2$ has more semantic information than $F_1$, and vice versa.

\subsubsection{Feature Similarity}
\label{sec:3.2.2}

\par Linear Centered Kernel Alignment (CKA) \cite{kornblith2019similarity} is a similarity index that can detect meaningful similarities between high-dimensional features. It also has great flexibility in measuring the feature similarity between two inputs with different sizes. We use CKA to quantify the similarity among features in both unimodal and multi-modal models.

\subsubsection{Visual Validation}

\par Grad-CAM \cite{selvaraju2017grad} uses gradients of a target concept flowing into the final convolutional layer to generate a localization map highlighting crucial areas for predicting the concept. We utilize Grad-CAM to visualize the highlighted areas for each concept, allowing us to qualitatively verify the conclusions drawn from quantitative methods. For segmentation, we replace the logits by summing the activation values at each pixel labeled by the target concept in the output from the final convolutional layer \cite{vinogradova2020towards}.



\subsection{Dissecting cross-modal correlation}

To dissect cross-modal correlations, we leverage the proposed $SVar$ score to analyze semantic variance, while employing the CKA method to investigate feature differences and cross-level evolution disparities between RGB and depth modalities.
We evaluate the semantic variance between RGB and depth modalities by calculating $SVar(F_I^S;F_D^S)$ and $SVar(F_I^J;F_D^J)$. To further explore the differences among features of the two modalities and their layers, we employ CKA across modalities and layers, as illustrated in Figure~\ref{fig:framework interpretation} (b).


\subsection{Dissecting the fusion process}

 
To reveal the inter-modal complementary gains achieved through fusion, we measure the semantic variance between separately-trained unimodal features and the joint RGB-D feature by calculating $SVar(F_C^J;F_I^S)$ and $SVar(F_C^J;F_D^S)$. 

To evaluate the semantic gains resulting from the fusion layer, we calculate the semantic variance between the concatenation of separately-trained unimodal features and the concatenation of the jointly-trained counterparts. We obtain $Cat(F_I^S, F_D^S)$ and $Cat(F_I^J, F_D^J)$ by concatenating the separately-trained features and jointly-trained features, respectively, where $Cat(\cdot)$ denotes channel-wise concatenation. The semantic gains from $Cat(F_I^S, F_D^S)$ to $Cat(F_I^J, F_D^J)$ are given by $SVar(Cat(F_I^J, F_D^J);Cat(F_I^S, F_D^S))$.

To assess the semantic gains resulting from joint training, we evaluate the semantic variance between the separately-trained unimodal features and jointly-trained counterparts by calculating $SVar(F_I^J;F_I^S)$ and $SVar(F_D^J;F_D^S)$.

\begin{figure*}[ht]
    \centering

    \includegraphics[width=0.45\linewidth]{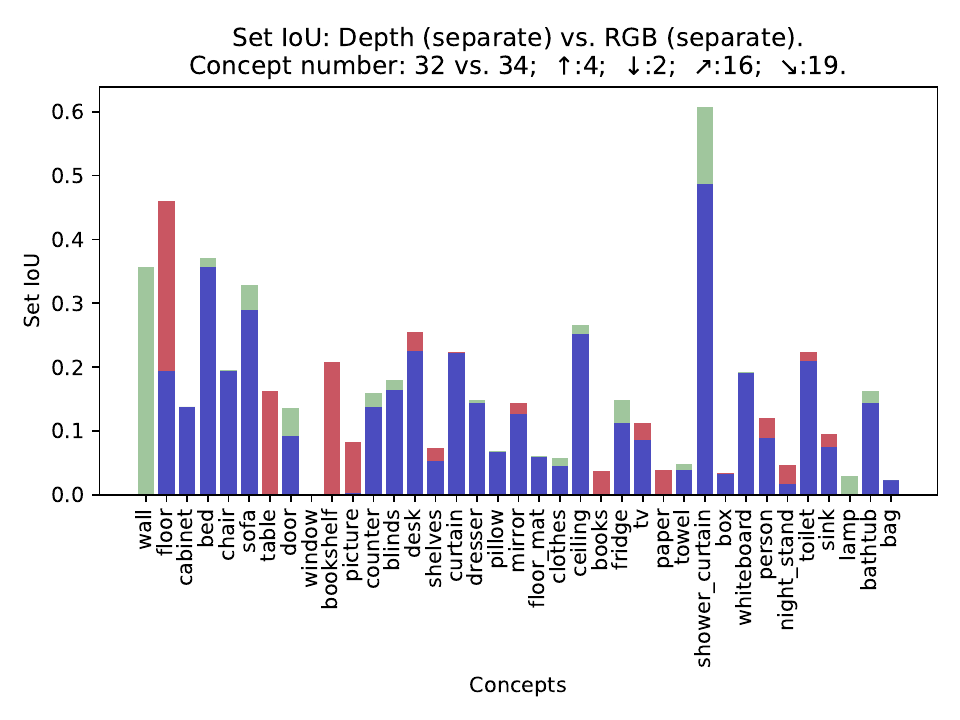}
    \hfill
    \includegraphics[width=0.45\linewidth]{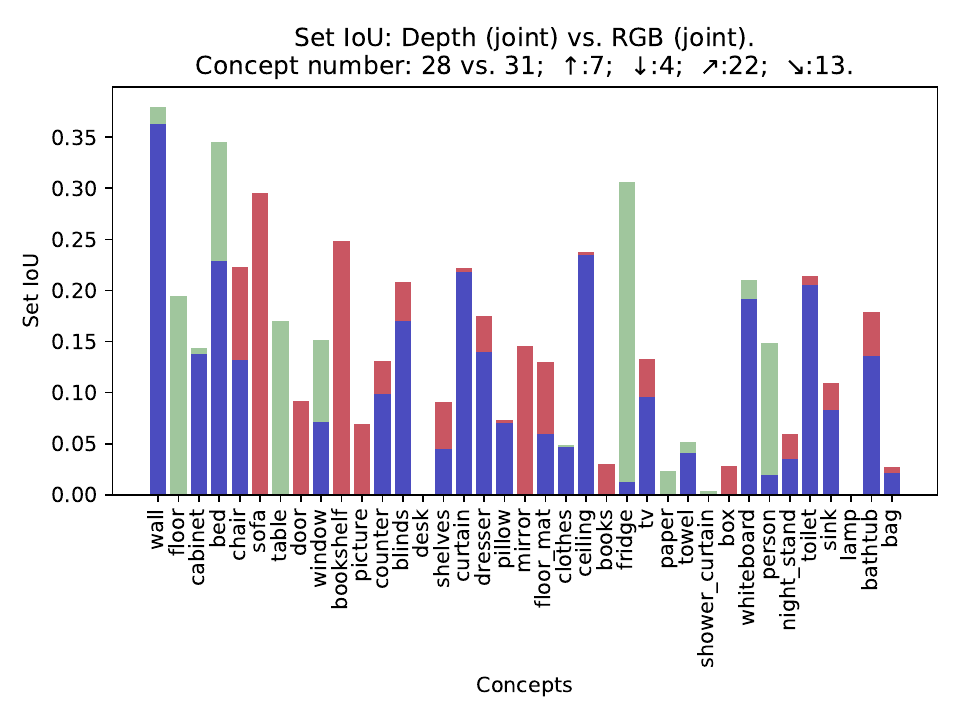}
    \vspace{-10pt}
    \caption{Comparison the Set IoU between depth and RGB. \textbf{Left:} separate training. \textbf{Right:} joint training.}
    \label{fig:depth(sep) vs rgb(sep)}
\end{figure*}

\section{Experiments}
\label{sec:exper}

\subsection{Implementation details}
\label{sec:imp det}
\noindent \textbf{Dataset.} We use the SUN-RGBD dataset \cite{song2015sun}, which contains 37 classes and 10,335 images for training and evaluation.
We sample 500 RGB-D pairs from the testing set to serve as the reference ``probe" dataset to train Net2Vec. 

\noindent \textbf{Models.} The network architectures we interpret include a simple unimodal model and an RGB-D multi-modal model. The unimodal model consists of a ResNet50 \cite{he2016deep} as encoder and a simple decoder. The multi-modal model uses two ResNet50 encoders to encode the input from each modality and concatenates the feature maps from both encoders. It then reduces the dimension of the concatenated output using a $1 \times 1$ convolutional layer as a dimensional reduction layer to obtain the final feature map that serves as input to the same decoder used in the unimodal model.

\noindent \textbf{Training settings.} All models are trained on the training set of the SUN-RGBD dataset, except for Net2Vec which is trained for 30 epochs on the reference “probe” dataset. 
All images and ground truth labels are resized to a spatial size of 480 $\times$ 640 pixels. We set the maximum number of epochs to 500 and use binary cross-entropy to supervise the training process. We use an SGD optimizer with an initial learning rate of 0.002, momentum of 0.9 and weight decay of 0.004. The batch size is 10 when training on one NVIDIA 3090 GPU and the learning rate is multiplied by 0.8 every 100 iterations.

\subsection{Modality specialty and discrepancy}




\noindent \textbf{RGB and depth specialize in different semantics.}
As demonstrated in previous multi-modal fusion architectures, combining high-level semantic cues in RGB-D pairs always leads to noticeable improvements. Therefore, the first question we want to investigate is how the two modalities complement each other in the semantic distribution. Figure~\ref{fig:depth(sep) vs rgb(sep)} illustrates the difference in IoU distributions between RGB and depth. To clearly show the difference, we present their Set IoU distributions together. In Figure~\ref{fig:depth(sep) vs rgb(sep)}, we consider the Set IoU of depth as the baseline, denoted by blue bars. The red and green parts represent the increased and decreased Set IoU values in RGB compared to the baseline, respectively. The $\uparrow$ and $\downarrow$ symbols denote newly emergent and vanished concepts, while $\nearrow$ and $\searrow$ denote concepts with increased and decreased Set IoU, respectively.

\begin{figure}[htbp]
    \centering
   \includegraphics[width=\linewidth]{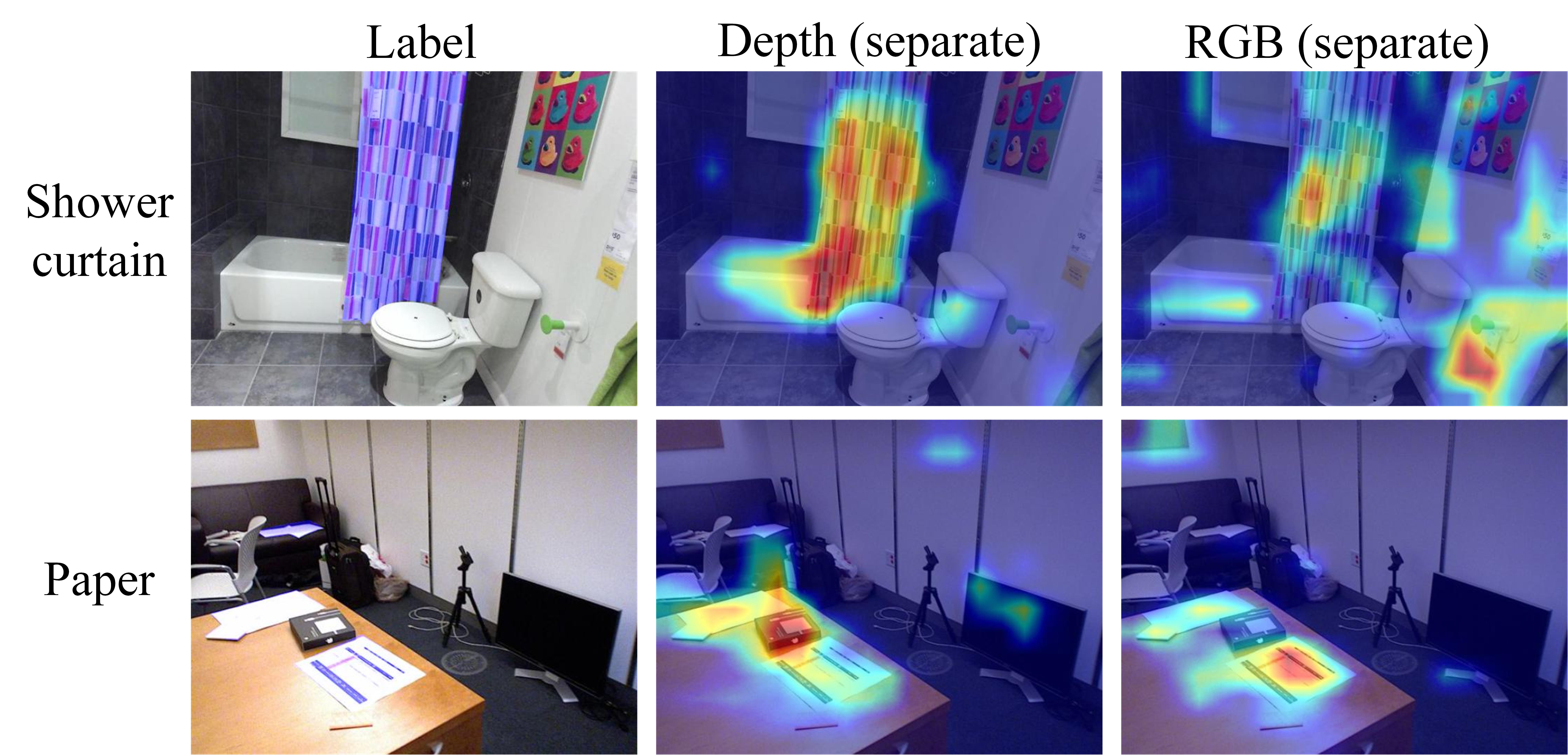}
   \caption{Visualization of the Grad-Cam from unimodal streams to show the cross-modal discrepancy.}
\label{fig:separate visualize}
\end{figure}

\begin{figure*}[htbp]
    \centering
    \subfloat[RGB]{
    \includegraphics[width=0.3\linewidth]{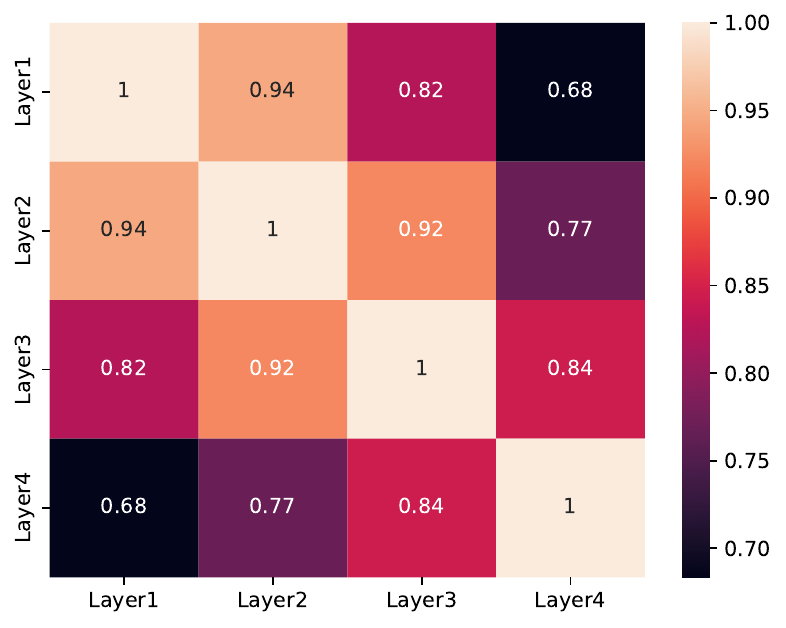}
    }
    \hfill
    \subfloat[Depth]{
    \includegraphics[width=0.3\linewidth]{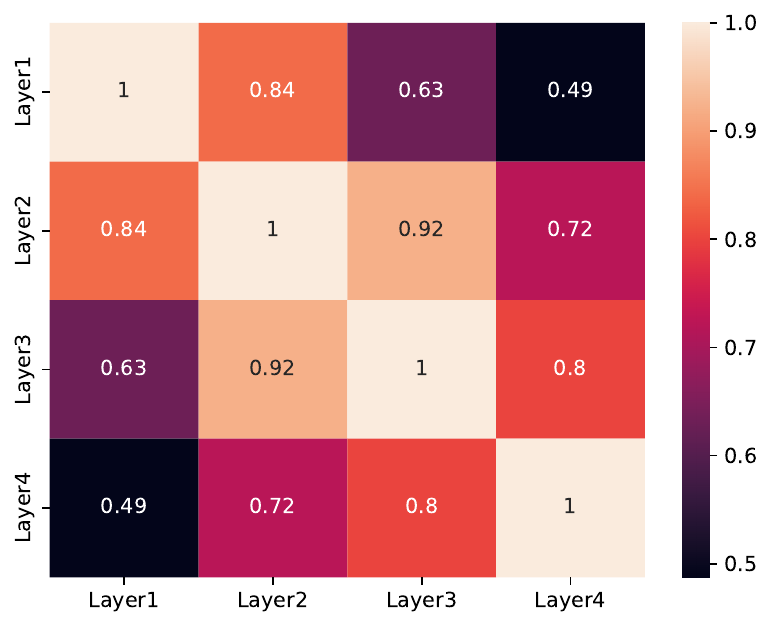}
    }
    \hfill
    \subfloat[Cross-modal CKA]{
    \includegraphics[width=0.3\linewidth]{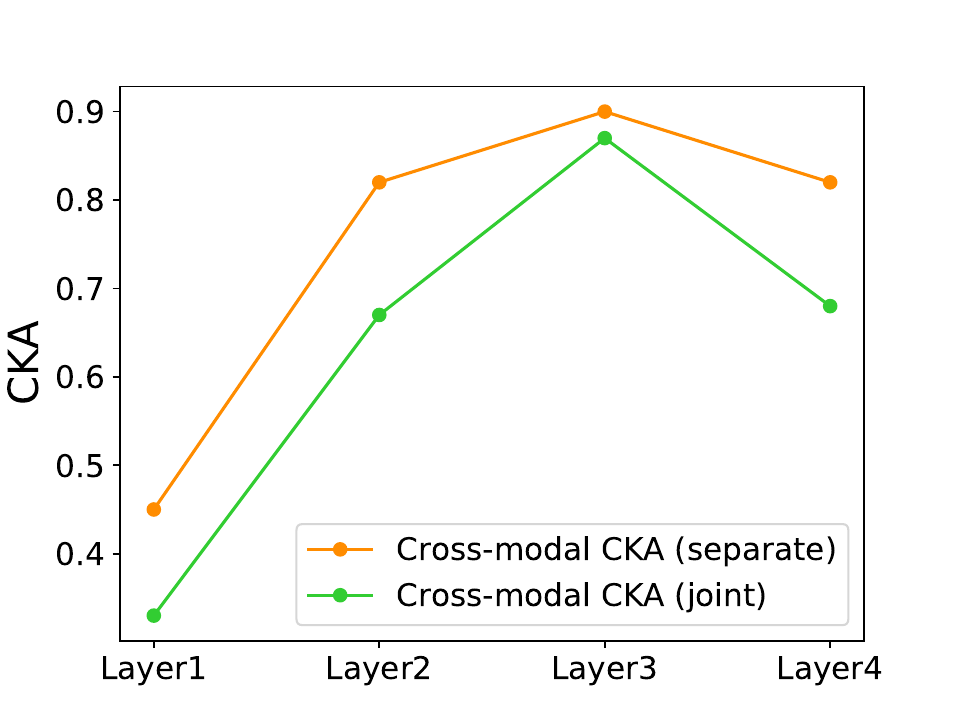}
    }
   \caption{Cross-level CKA in RGB (a) and depth (b), measured in unimodal networks. (c) Cross-modal CKA in each layer and its change with joint training.}
\label{fig:Cross-level CKA}
\end{figure*}

\par By comparing the unimodal Set IoU in Figure~\ref{fig:depth(sep) vs rgb(sep)}, we found that \textbf{RGB generally performs better in understanding more concepts than depth, and the two modalities specialize in different classes}. Specifically, RGB performs better in classes with abundant appearance/texture cues, such as the floor and bookshelf, while depth specializes in classes with large depth variance (e.g., wall) or rich geometry cues (e.g., bed and sofa). The visualization in Figure~\ref{fig:separate visualize} also verifies this cross-modal discrepancy. These quantitative and qualitative comparisons clearly demonstrate the semantic complementarity between RGB and depth modalities, which is well in accordance with their sensor characteristics.


\noindent \textbf{The cross-modal gap varies with layers.} The orange line in Figure~\ref{fig:Cross-level CKA}(c) shows the cross-modal CKA in each layer. We find that the cross-modal gap varies in different layers. The first level holds a much greater gap than other deeper layers, and \textbf{the cross-modal similarity generally increases with the network going deeper}. This variance inspires us to pay more attention to train the shallow layers when introducing a new modality. Also, we can refer to this variance to control the strictness when paying cross-modal consistency constraints for cross-modal transfer learning.

\noindent \textbf{RGB and depth hold different cross-level evolutions.}
We are also interested in whether the cross-modal disparity exists in the model optimization process. Figure~\ref{fig:Cross-level CKA} reports the cross-modal discrepancy in terms of the cross-layer similarity, indicating that depth holds a larger cross-layer disparity than RGB. This difference indicates that \textbf{the two modalities have distinct feature abstraction schedules and motivates us to customize a network with befitting depth and kernel size for new modalities}.

\begin{figure*}[htbp]
    \centering
    \includegraphics[width=\linewidth]{figures/4fig2.jpg}  
    \vspace{-12pt}
   \caption{(a) Comparison between the separately-trained unimodal features and jointly-trained counterparts for each modality. (b) Comparison between the jointly-trained unimodal features and the joint RGB-D feature.}
\label{fig:sep vs joint}
\end{figure*}
\begin{table}[htbp]
\centering
\caption{Semantic variance ($SVar$) for varying pairs.}
\label{tab:1}
\begin{tabular}{c|l|c}
\hline
Group                   & Layers                              & Semantic variance  \\
\hline
\multirow{2}{*}{1}      & Depth (separate) vs. RGB (separete) & +118.41            \\
                        & Depth (joint) vs. RGB (joint)       & +165.72            \\
\hline
\multirow{2}{*}{2}      & Depth (separate) vs. Depth (joint)  & -50.45            \\  
                        & RGB (separate) vs. RGB (joint)      & -156.67           \\
\hline
\multirow{2}{*}{3}      & Depth (separate) vs. RGB-D (joint)  & +194.08           \\
                        & RGB (separate) vs. RGB-D (joint)    & +52.76             \\
\hline
\multirow{2}{*}{4}      & Depth (joint) vs. RGB-D (joint)     & +324.79            \\
                        & RGB (joint) vs. RGB-D (joint)       & +203.65            \\
\hline
5                       & Cat (separate) vs. Cat (joint)     & +72.71             \\
\hline
\end{tabular}
\end{table}

\begin{figure}[H]
\begin{center}
    \includegraphics[width=\linewidth]{figures/joint_visualize.jpg}
\end{center}
   \vspace{-5pt}
   \caption{Visualization of the Grad-Cam to show the cross-modal fusion gains when trained jointly.}
\label{fig:1}
\end{figure}
\subsection{Fusion mechanism}
\noindent \textbf{What and by how much do two modalities complement?} Figure~\ref{fig:sep vs joint}(b) shows that the fusion layer improves the Set IoU of a large majority of concepts in each modality and identifies noticeable concepts that were previously undetected. This indicates that \textbf{the fusion layer effectively combines the semantics of each modality to create a comprehensive representation.} Figure~\ref{fig:1} provides visual examples of the benefits of fusion, where the fusion layer integrates the semantics from each modality to make comprehensive joint inference. In Table~\ref{tab:1}, group 4 shows that compared to the joint feature, RGB has a smaller semantic gain than depth, which suggests that \textbf{RGB carries more semantic cues and is the dominant modality in the multi-modal system.}




\begin{figure}[H]
    \centering
    \includegraphics[width=\linewidth]{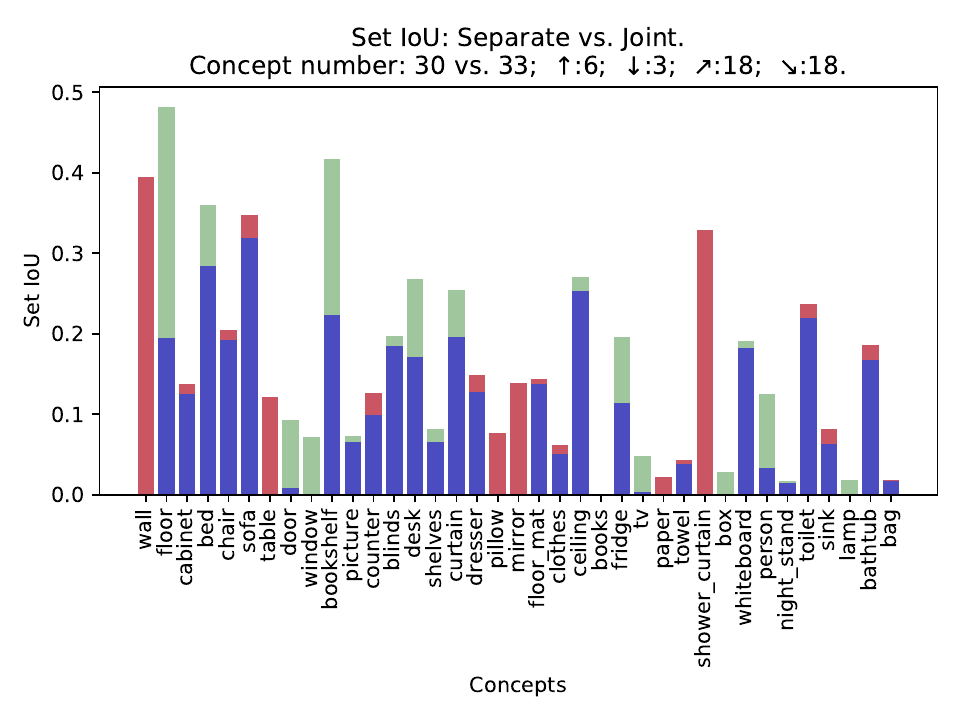}
    \vspace{-20pt}
    \caption{Comparison between the concatenation of separately-trained features and the joint RGB-D feature.}
    \label{fig:0}
\end{figure}

To demonstrate the additional benefits of joint training, we concatenate the separately-trained unimodal features and the jointly-trained counterparts, respectively, and denote them as ``Separate''($Cat(F_I^S, F_D^S)$) and ``Join''($Cat(F_I^J, F_D^J)$) in Figure~\ref{fig:0}. Comparing their Set IoU distributions, we find that training two modalities jointly with a fusion layer achieves substantial gains. In Table~\ref{tab:1}, group 5 shows that joint training introduces +72.71 additional semantic gains over directly concatenating separately-trained unimodal features. Specifically, the Set IoU of joint training covers more concepts as \textbf{the fusion layer can integrate cross-modal semantic cues to identify new classes.}

\noindent \textbf{Complementarity logic: Consistency or Specificity?} While each modality acquires new semantic cues from the other, they do not optimize towards a similar feature. Figure~\ref{fig:depth(sep) vs rgb(sep)} and group 1 in Table~\ref{tab:1} demonstrate that cross-modal semantic discrepancy is increased with fusion (from +118.41 to +165.72), despite some consistent highlighting of concepts by both modalities. Furthermore, Figure~\ref{fig:Cross-level CKA}(c) shows a consistent decrease in cross-modal CKA values in each layer after fusion, indicating a drop in cross-modal alignment. This suggests that \textbf{the fusion layer may redistribute the semantic recognition responsibility of the modalities based on their specialties, resulting in significant disturbance in the semantic distribution of each modality.} \textbf{Therefore, the multi-modal fusion logic appears to be a hybrid approach with simultaneous cross-modal competition and collaboration.} Simply minimizing or maximizing their consistency, as done in previous works, may lead to suboptimal results.

\begin{figure*}[htbp]
    \centering
    \subfloat[]{
    \includegraphics[width=0.45\linewidth]{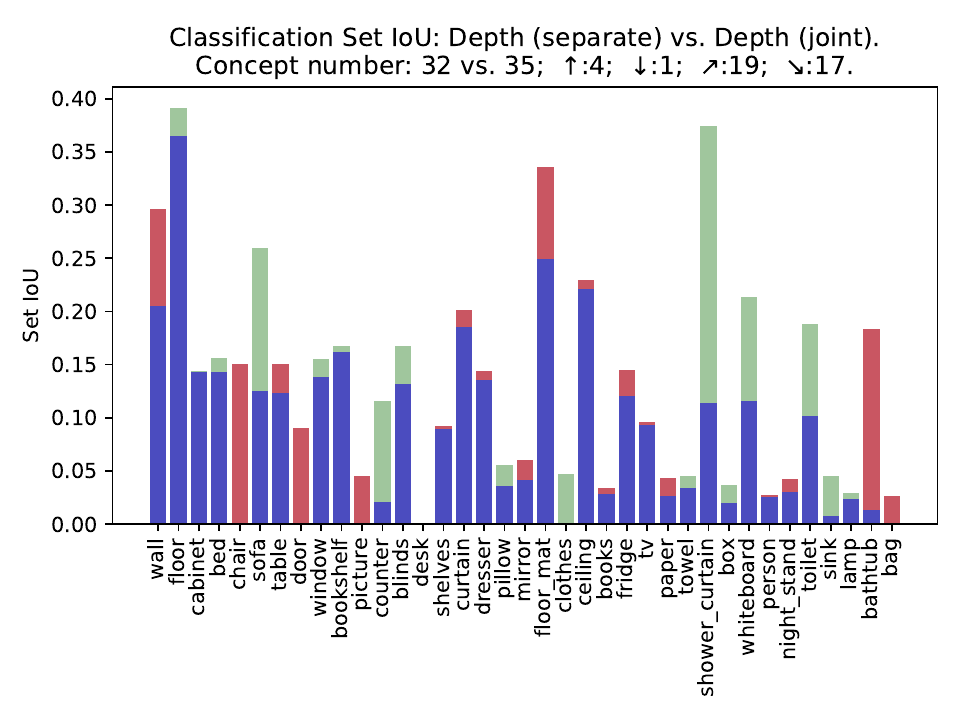}
    }
    \hfill
    \subfloat[]{
    \includegraphics[width=0.45\linewidth]{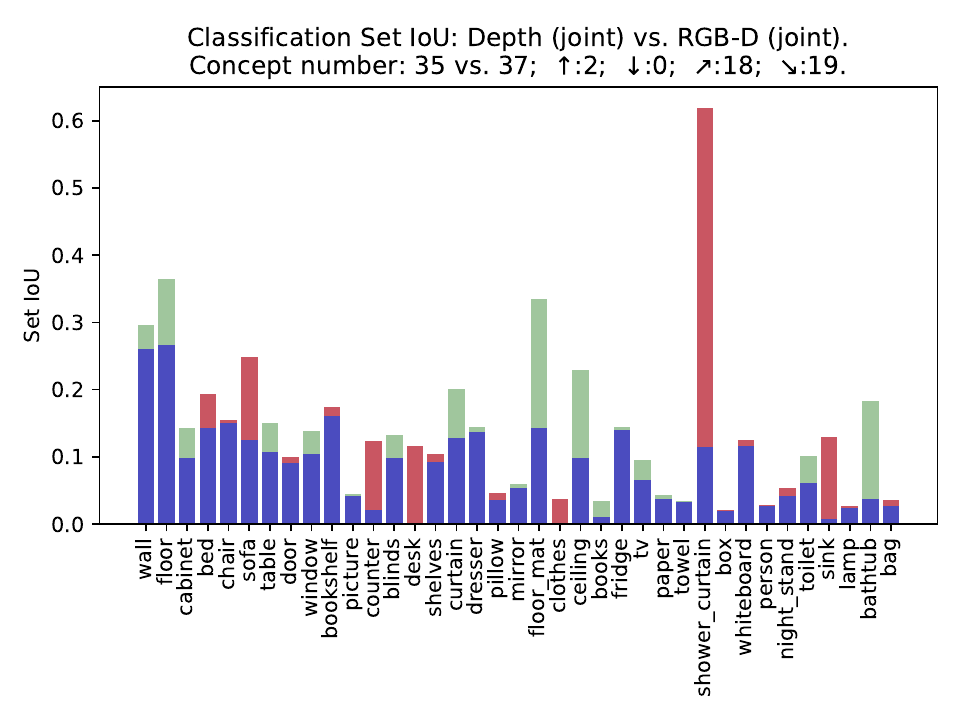}
    }
   \caption{(a) Comparison between unimodal streams and the jointly trained counterpart for depth modality in classification task. (b) Comparison between the depth feature trained jointly and the joint RGB-D feature in classification task.}
\label{fig:cls sep vs joint}
\end{figure*}

\noindent \textbf{How does each modality evolve with the fusion process?} Apart from the fusion gains, we are also interested in how each modality evolves with the fusion process compared to the counterparts trained separately. Figure~\ref{fig:sep vs joint}(a) shows the semantic variance for each modality after joint training, where both RGB and depth experience noticeable semantic variance. Specifically, we find that, with the cross-modal fusion process, each modality activates new concepts (e.g., in RGB and in depth). Surprisingly, as shown in group 2 in Table~\ref{tab:1}, the overall semantic variance for both modalities is negative, indicating that the overall semantic information for each modality is decreased. By analyzing the IoU variance in Figure~\ref{fig:sep vs joint}(a), we find that a considerable number of concepts experience a severe change in IoU values in each modality, and several concepts in each modality vanish after joint training. Comparing this change to the semantic variance between the jointly-trained RGB/depth and RGB-D joint features in Figure~\ref{fig:sep vs joint}(b), we observe that those vanished/decreased concepts in the unimodal Set IoU are re-activated/enhanced in the fused RGB-D one (e.g., floor, sofa, and table). Hence, we argue that \textbf{the semantic decrease from separately training to jointly training for each modality is an intermediate state resulting from the dynamic competitive collaboration mechanism in the fusion layer.} \textbf{In addition to exchanging inter-modal cues to cooperatively highlight some concepts, the fusion layer also tries to force each modality to be responsible for different concepts.} Consequently, for each modality shown in group 3 and 4 in Table~\ref{tab:1}, the jointly-trained stream exhibits larger semantic gains than the separately-trained one, compared to the fused RGB-D features.     
\par Moreover, the emergent concepts are precisely the well-learned ones in the paired modality, suggesting that \textbf{cross-modal complementarity not only lies in joint inference but also motivates significant mutual promotion for each modality in the feature learning stage.} We also find that \textbf{depth emerges more new concepts than RGB, partly verifying the modality imbalance, and that depth is the weaker modality.}
\subsection{Generalize to other tasks}
Our multi-modal dissection framework can be easily generalized to other tasks by adapting the inference head in the decoder.

\begin{table}[htbp]
\centering
\caption{Semantic variance ($SVar$) for varying pairs in classification task.}
\label{tab:2}
\begin{tabular}{c|l|c}
\hline
Group                   & Layers                              & Semantic variance  \\
\hline
\multirow{2}{*}{1}      & Depth (joint) vs. RGB-D (joint)     & +15.98            \\
                        & RGB (joint) vs. RGB-D (joint)       & +23.06            \\
\hline
2                       & Cat (separate) vs. Cat (joint)     & +30.13            \\
\hline
\end{tabular}
\end{table}

\noindent \textbf{Generalize to classification.}
In the classification task, we modify our models by changing the decoder to a classification head. As shown in Figure~\ref{fig:cls sep vs joint}, we find that the dynamic competitive collaboration mechanism in the fusion process also occurs in the classification task. Moreover, Table~\ref{tab:2} demonstrates that jointly training can bring semantic gains in the classification task, as evidenced by groups 1 and 2. However, the semantic variances introduced by the fusion classification model is smaller than that in the segmentation task because segmentation requires more semantic information for each region.

\subsection{Applications of our findings}
Based on our findings that \textit{the multi-modal fusion logic appears to be a hybrid approach with simultaneous cross-modal competition and collaboration.} We design a simple strategy with negligible parameters, which is called \textit{Bidirectional Mutual Constraint (BMC)}. In this strategy, for a multi-modal fusion model, we split the features from each modality into two halves before their multi-modal fusion layers. One half of the features undergoes cross-modal mutual information minimization, while the other half undergoes cross-modal mutual information maximization. This strategy aims to promote both competition and collaboration between the modalities.

\begin{table}[htbp]
\centering
\caption{Results of CMINet and variant using our strategy on two RGB-D saliency detection datasets. By replacing its  minimization constraint with ours, we achieve noticeable gains.}
\label{tab:3}
\begin{tabular}{lc|c|c}
  \hline
    & Metric &
   CMINet & CMINet+BMC \\
   \hline
  \multirow{4}{*}{\rotatebox{90}{\textit{SIP}}}
    & $S_{\alpha}\uparrow$    & 0.885 & \textbf{0.897}  \\
    & $F_{\beta}\uparrow$     & 0.880 & \textbf{0.892}  \\
    & $E_{\xi}\uparrow$       & 0.919 & \textbf{0.932}  \\
    & $\mathcal{M}\downarrow$ & 0.048 & \textbf{0.042}  \\ \hline
  \multirow{4}{*}{\rotatebox{90}{\textit{LFSD}}}
    & $S_{\alpha}\uparrow$    & 0.867 & \textbf{0.875}  \\
    & $F_{\beta}\uparrow$     & 0.848 & \textbf{0.862}  \\
    & $E_{\xi}\uparrow$       & 0.890 & \textbf{0.904}  \\
    & $\mathcal{M}\downarrow$ & 0.070 & \textbf{0.064}  \\ \hline
  \end{tabular}
\end{table}

\begin{table}[htbp]
\centering
\caption{Results of ACNet and variant using our strategy for semantic segmentation on NYU-Depth V2 dataset.}
\label{tab:4}
\begin{tabular}{c|c|c}
  \hline
    Metric &
   ACNet & ACNet+BMC \\
   \hline
    mIoU & 0.419 & \textbf{0.432}  \\ \hline
\end{tabular}
\end{table}

\begin{table}[H]
\centering
\caption{Results of AMC and variant using our strategy for visual grounding on RefCOCO and RefCOCO+.}
\label{tab:5}
\begin{tabular}{lc|c|c}
  \hline
    & Dataset &
   AMC & AMC+BMC \\
   \hline
  \multirow{2}{*}{\textit{RefCOCO}}
    & test A       & 77.13 & \textbf{78.36}  \\
    & test B & 64.89 & \textbf{66.50}  \\ \hline
    \multirow{2}{*}{\textit{RefCOCO+}}
    & test A    & 80.34 & \textbf{81.42}  \\
    & test B    & 64.55 & \textbf{66.76}  \\ \hline
\end{tabular}
\end{table}

We apply our BMC strategy to three state-of-the-art models on three different tasks: ACNet~\cite{hu2019acnet} (Semantic Segmentation), CMINet~\cite{MIM} (Salient Object Detection), and AMC~\cite{yang2023improving} (Visual Grounding). Specifically, for the BMC strategy applied to ACNet and CMINet, we utilize the mutual information evaluation method proposed in CMINet to obtain the cross-modal mutual information. In the case of AMC, we use the contrastive learning approach. Note that, we only split the original features into two halves without introducing additional parameters. The other experimental settings remain the same, and the results are summarized in Table~\ref{tab:3}, Table~\ref{tab:4} and Table~\ref{tab:5}. Our strategy averagely improves ACNet by 1.3\% in mIoU, improves CMINet by 1.3\% in F-measure, and improves AMC by 1.5\% in pointing game accuracy.
We find that our strategy achieves consistent and large improvement, denoting the importance and generalization of our findings and the proposed analysis framework.
It is worth noting that CMINet uses a strategy of minimizing mutual information, which is clearly inferior to our approach that simultaneously encourages both the minimization and maximization of mutual information. This also validates the correctness and practical value of our conclusions.

\section{Conclusion}
\label{sec:conc}

In this work, we make the first attempt to construct an analytical framework for dissecting RGB-D fusion. By designing a metric to quantify semantic variance, we provide an explicit dissection of inter-modal complementarity, multi-modal complementing logic and unimodal promotion with fusion. Our discoveries can serve as a basis for designing or diagnosing multi-modal visual models. Experiments also verify the efficacy and generalization of proposed fusion strategy derived from our findings.




\ifCLASSOPTIONcaptionsoff
  \newpage
\fi

{
\bibliographystyle{IEEEtran}
\bibliography{IEEEabrv,mybibfile}
}

\vfill

%




\end{document}